\documentclass[runningheads]{llncs}
\usepackage[T1]{fontenc}
\usepackage{graphicx}
\usepackage{epsfig}
\usepackage{amsmath}
\usepackage{mathtools}
\usepackage{amsfonts}       

\usepackage[nameinlink,capitalize]{cleveref}
\usepackage{booktabs}       
\usepackage{subfigure}
\usepackage{algorithm}
\usepackage{algpseudocode}
\usepackage{soul}
\usepackage{ragged2e}
\usepackage{xcolor}
\usepackage{microtype}

\newcommand{\tp}{{\text{\scriptsize\sffamily T}}}

\begin{document}
\title{What are You Looking at? Modality Contribution in Multimodal Medical Deep Learning}
\titlerunning{What are You Looking at?}
%
\author{Christian Gapp\inst{1,2}\orcidID{0000-0002-4520-298X} \and
Elias Tappeiner\inst{1}\orcidID{0000-0003-1034-8361} \and
Martin Welk\inst{1}\orcidID{0000-0002-6268-7050} \and 
Karl Fritscher\inst{2}\orcidID{0000-0003-2593-6203} \and
Elke R. Gizewski\inst{3}\orcidID{0000-0001-6859-8377} \and
Rainer Schubert\inst{1}\orcidID{0000-0002-8026-7500} }
\authorrunning{C. Gapp et al.}
%
\institute{Institute of Biomedical Image Analysis\\
	UMIT TIROL -- Private University for Health Sciences and Health Technology,\\	Eduard-Wallnöfer-Zentrum 1, 6060 Hall in Tirol, Austria \and
VASCage -- Centre on Clinical Stroke Research, 6020 Innsbruck, Austria \and
Department of Radiology, Medical University of Innsbruck, 6020 Innsbruck, Austria\\
\email{\{christian.gapp, elias.tappeiner, martin.welk, rainer.schubert\}@umit-tirol.at,\\karl.fritscher@vascage.at, elke.gizewski@i-med.ac.at}}
%
\maketitle              

\begin{abstract}

\emph{Purpose}
High dimensional, multimodal data can nowadays be analyzed by huge deep neural networks with little effort. Several fusion methods for bringing together different modalities have been developed. Given the prevalence of high-dimensional, multimodal patient data in medicine, the development of multimodal models marks a significant advancement. However, how these models process information from individual sources in detail is still underexplored.\newline
\emph{Methods}
To this end, we implemented an occlusion-based modality contribution method that is both model- and performance-agnostic. This method quantitatively measures the importance of each modality in the dataset for the model to fulfill its task.
We applied our method to three different multimodal medical problems for experimental purposes.\newline
\emph{Results}
Herein we found that some networks have modality preferences that tend to unimodal collapses, while some datasets are imbalanced from the ground up. Moreover, we provide fine-grained quantitative and visual attribute importance for each modality.\newline
\emph{Conclusion}
Our metric offers valuable insights that can support the advancement of multimodal model development and dataset creation. By introducing this method, we contribute to the growing field of interpretability in deep learning for multimodal research. This approach helps to facilitate the integration of multimodal AI into clinical practice. Our code is publicly available at 
{\tt{{https://github.com/ChristianGappGit/MC\_MMD}}}.

\end{abstract}

\section{Introduction}

Multimodal datasets, especially in the field of medicine, are becoming ever larger and more present. Thus, over the past few years, multiple types of multimodal fusion methods, such as those summarized in \cite{MultiModalTransformersSurvey}, have been developed to process high-dimensional multimodal data. It is of utmost interest to develop interpretability methods that can explain the deep learning based model's behavior on such multimodal datasets for solving a specific task.

Clinical applications that use multimodal data include cancer prognosis prediction \cite{cancers14133215}, cardiovascular risk assessment \cite{cardio_multimodal_learning}, Parkinson’s diagnosis \cite{parkinson_disease_pred}, or, for instance, diabetic retinopathy classification \cite{diabetic_retinopathie_classification}.
Interpretability methods for the implemented models have the potential to enhance credibility and thereby accelerate the integration of multimodal AI into clinical practice more broadly -- \emph{"a bridge to trust between technology and human care"}~\cite{Beger2024}.

Interpretability methods, such as GradCAM \cite{GradCAM} and Occlusion Sensitivity \cite{OccSens}, have been created for single modality based networks. However, for multimodal data and models these methods are yet very underexplored. Most existing methods lack quantification of modality importance and thus inhibit comparability between models and datasets. Some existing modality importance methods either depend on the model’s performance \cite{gat2021perceptualscoredatamodalities,hu2022shapeunifiedapproachevaluate} or on the architecture itself \cite{coffeBean,MCI_SHAP}. Others, such as attention and gradient based methods, are not sufficient to measure a modality's importance in multimodal datasets \cite{jain-wallace-2019-attention}.

To the best of our knowledge, we close a gap by creating a performance and model agnostic (black-box) metric to measure the modality contribution in multimodal tasks and are the first to test it on medical datasets: one image--text dataset (2D Chest X-Rays + clinical report) from Open I \cite{OpenI_dataset}, one image--tabular dataset (2D color ophthalmological images + patient information) BRSET \cite{BRSET} and another image--tabular dataset (3D head and neck CT + patient information), viz. Hecktor~22 \cite{Hecktor22}, published within a MICCAI grand challenge 2022.

With our metric it is possible to detect unimodal collapses, i.e. whether the model focuses extensively or even exclusively on one modality to solve a problem (e.g. \cite{madhyastha-etal-2018-defoiling}).
Furthermore, architectures can be compared regarding their ability to process different modalities within one dataset.


\section{Related Work}\label{sec:RelatedWork}
The survey \cite{Acosta2022} summarizes the potential of using multimodal data in biomedical and medical tasks for clinical practice. Multimodal medical data has been successfully used in a range of tasks: the prediction of cancer prognoses \cite{cancers14133215}, cardiovascular risk stratification using electronic health records and imaging \cite{cardio_multimodal_learning}, early diagnosis of Parkinson’s disease from mobile-derived sensor data \cite{parkinson_disease_pred}, diabetic retinopathy severity classification from combined imaging modalities \cite{diabetic_retinopathie_classification}, etc. Ngiam et al. (2011) pioneered multimodal deep learning by showing that integrating data from different modalities, such as audio and video, can enhance feature learning and representation quality \cite{ngiam2011multimodal}.
As understanding the behavior of models on multimodal data remains challenging, interpretability methods are essential for building trust in multimodal AI systems \cite{Beger2024}.

In general, in order to measure a modality's contribution to a task, performance agnostic or performance dependent metrics, model agnostic (black box model), or non-model agnostic (white box) metrics are distinguished.
In \cite{Jin_2022} the authors found that attention based explainability methods can not measure single feature importance adequately. Hence methods for investigating modality contributions based on these non-model agnostic (attention) methods will suffer from the same inability to select important features. Therefore black-box based modality importance methods are actually of utmost interest.

Several feature importance methods are widely used. Permutation importance, proposed by Breiman~\cite{breiman2001random} and extended by Fisher et al.~\cite{fisher2019all}, measures performance drop when features are shuffled. Occlusion-based techniques~\cite{OccSens}, like mean replacement or feature removal, are common for tabular data. Model-agnostic methods include SHAP~\cite{Shapley}, which uses Shapley values, and LIME~\cite{ribeiro2016lime}, which fits local surrogate models.

In multimodal settings, importance scores based on performance metrics have been proposed in~\cite{gat2021perceptualscoredatamodalities,hu2022shapeunifiedapproachevaluate}. In contrast, \cite{coffeBean} uses a Shapley-based~\cite{Shapley} approach that is performance-independent but requires access to model architecture. A non-model agnostic/\-task metric for modality selection is introduced in~\cite{MCI_SHAP}.

Within this work we develop a multimodal importance metric that is both model and performance agnostic.

\section{Methodology -- Interpretability}\label{sec:Interpretability}
In the following we present our occlusion-based modality contribution method -- a quantitative measure of the importance of modalities in multimodal datasets processed by multimodal neural networks.

\subsection{Modality Contribution $m_i$}
We define the metric $m_i \in [0,1]$ as a quantification for the contribution (importance) of the modality $i$ ($i=1,\dots, n$) in the dataset with $n$ modalities for a specific problem that was solved with a specific multimodal deep learning model. The sum over all modality contributions $\sum_{i=1}^{n} m_i= 1$ remains constant.

Let $\vec{p}^k_0$ be the model's output vector for the sample $k = 1, \dots, N$ in the dataset with $N$ samples and $\vec{x}^k_i$ the input vector containing the features of sample $k$ for modality $i$. Then one can compute a modality specific output $\vec{p}^k_i$ for sample $k$ by manipulating $\vec{x}^k_i$ and storing the absolute difference $\vec{d}^k_i = \lvert \vec{p}^k_0 - \vec{p}^k_i\rvert$. Repeating this manipulation over all samples and averaging the output differences will result in $\vec{d}_i = \sum_{k=1}^{N}\vec{d}_i^k / N$. After this is done for all modalities, we finally can compute 
\begin{equation}
	m_i = \frac{\vec{1}^\tp\vec{d}_i}{\sum_{j=1}^n \vec{1}^\tp\vec{d}_j + \epsilon},
\end{equation}
with $\epsilon \approx 10^{-7}$.
Herein the manipulation of the input vector $\vec{x}^k_i$ is the key process. It can be realized in different ways. One can mask the whole vector (i.e. the whole modality) by replacing all entries in $\vec{x}^k_i$ with zeros or with the mean of the modality specific entries in all samples, for instance. Our method works with higher resolution as we mask parts of $\vec{x}^k_i$ and repeat the forwarding process of the manipulated data to the model more often for one sample and one modality. 

We split the vector $\vec{x}_i^k$ into $h_i$ parts and store the intermediate output distances $\vec{d}_{i,l}^k = \lvert \vec{p}_{0}^k-\vec{p}_{i,l}^k\rvert$, with $l = 0, \dots, h_i-1$ before obtaining $\vec{d}_i^k = \sum_{l=0}^{h_i-1}\vec{d}_{i,l}^k$. 
Therein $h_i$ is a modality specific hyper-parameter. When processing tabular data, we can easily mask each entry in $\vec{x}^k_i$ itself ($h_i  = \text{length}(\vec{x}_i^k)$). For the vision modality we mask pixel or voxel patches in order to get interpretable results and thereby keep computation time limited due to large vision input ($h_i = \prod_{d = 0}^{D-1}\text{img\_shape}[d] \, / \, \text{patch\_shape}[d]$, with number of image dimensions $D$).

The only criterion for $h_i$ is $h_i < h_{i,{\text{max}}}$, with upper limit $h_{i,{\text{max}}}$, where the model detects no significant differences in the output for the slightly masked input. Assuming we mask every pixel (2D) or voxel (3D), the model would not be affected sufficiently. The contribution of the vision modality would be underestimated. As long as one patch can occlude significant information, which is normally already the case for small masks too, $h_i$ is small enough to ensure adequate modality dependent dynamic in the model. However, choosing smaller $h_i$, i.e. bigger patches, is straightforward and does not affect the estimation of the model contribution substantially. Moreover it is computationally more efficient to choose big patches.

\cref{alg:modality_contrib} summarizes the computation process of $m_i$ in the form of pseudo-code. 

\begin{algorithm}[t!]
	\caption{Computation of Modality Contribution $m_i$}\label{alg:modality_contrib}
	\begin{algorithmic}[1]
		\State $ \vec{d} \gets 0$
		\For{$i$ \textbf{in} $1$ \textbf{to} $n$}
		\State $\vec{d}_i \gets 0$
		\For{$k$ \textbf{in} $1$ \textbf{to} $N$}
		\State $\vec{x}^k \gets$ input\textsubscript{k}
		\State $\vec{p}_0^k \gets$ model($\vec{x}^k$)
		\State $\vec{d}_i^k \gets 0$
		\For{$l$ \textbf{in} $0$ \textbf{to} $h_i-1$}
		\State $\vec{x}_{i,l}^k \gets$ masked\_input\textsubscript{i,l,k}
		\State $\vec{p}_{i,l}^k \gets$ model($\vec{x}_{i,l}^k$)
		\State $\vec{d}_{i,l}^k \gets \lvert\vec{p}_{0}^k-\vec{p}_{i,l}^k\rvert$
		\State $\vec{d}_i^k \gets \vec{d}_i^k + \vec{d}_{i,l}^k$ 
		\EndFor
		\State $\vec{d}_i \gets \vec{d}_i + \vec{d}_i^k / N$
		\EndFor
		\State $\vec{d} \gets \vec{d} + \vec{d}_i$
		\EndFor
		\For{$i$ \textbf{in} $1$ \textbf{to} $n$}
		\State $m_i \gets \vec{1}^\tp\vec{d}_i / (\vec{1}^\tp \vec{d}+\epsilon)$
		\State $\vec{m} \gets [\vec{m},m_i]$
		\EndFor
	\end{algorithmic}
\end{algorithm}

\subsection{Modality-Specific Importance $mp_i^l$}
Independent of the modality contribution to the task itself, we provide a modality specific importance distribution on patches within one modality. 

The metric $mp_i^l \in [0,1]$ is defined as the contribution of a patch $l$ to the task relative to the other patches of the same modality $i$, with $l = 0, \dots, h_i-1$. We already computed the necessary factors $\vec{d}_{i,l}^k$, i.e. the distance between model output $\vec{p}_{i,l}^k$ with masked input $\vec{x}_{i,l}^k$ to plain output $\vec{p}_0^k$ with unmasked input $\vec{x}^k$. Thus we just have to store these factors, sum over all samples to get average distance $\vec{d}_{i,l} = \sum_{k=1}^{N}\vec{d}_{i,l}^k / N$ and finally compute
\begin{equation}
	mp_i^l = \frac{\vec{1}^\tp\vec{d}_{i,l}}{\sum_{j=0}^{h_i-1}\vec{1}^\tp\vec{d}_{i,j} + \epsilon}
\end{equation}
with $\epsilon \approx 10^{-7}$.
Note that $\sum_{l=0}^{h_i-1}mp_i^l = 1$, but $\sum_{i=1}^{n}mp_i^l \not= 1$, since $mp_i^l$ is modality-specific.

If we are interested in the contribution of the patch $p_i^l$ of modality $i$ to the overall task, we can simply weight $mp_i^l$ with the modality contribution $m_i$:
\begin{equation}
	mp_i^l \cdot m_i,
\end{equation}
fulfilling $\sum_{i=1}^{n}\sum_{l=0}^{h_i-1} mp_i^l \cdot m_i = 1$.

\subsection{Metric Properties}

\paragraph{Performance Independence}
Our metric is performance agnostic as we do compute the modality contribution to a specific task by measuring the dynamic it generates in the output. 

\paragraph{Normalization}
A modality's contribution to a specific task using a specific dataset is generated by normalizing the model's output dynamic over all modalities using all samples in the dataset. That enables a comparison between different tasks and architectures.

\paragraph{Applicability}
Since the computation of our metric is a \emph{black-box method}, it is applicable to every multimodal dataset and every model architecture with little effort.  

\section{Multimodal Medical Tasks}\label{sec:MultimodalMedicalTasks}
In order to test our modality contribution method, we first trained three multimodal medical tasks, viz. two classification problems and one regression problem. For this, the architectures given in \cref{tab:archs} were built. Details on the tasks together with trainings results are presented afterwards.

\setlength{\tabcolsep}{6pt}
\begin{table}[h!]
	\caption{Multimodal architectures used in our experiments.}
	\centering
	\begin{tabular}{llclclc}
	\toprule
	\textbf{architecture} &$\gets$ &\textbf{vision} &+ &\textbf{text/tabular} &+ &\textbf{fusion}\\
	\cmidrule{1-7}
	ViTLLaMAII &$\gets$ &ViT &+ &LLaMA~II &+ &Cross T.\\
	ResNetLLaMAII &$\gets$ &ResNet &+ &LLaMA~II &+ &Cross T.\\
	ViTMLP &$\gets$ &ViT &+ &MLP &+ &MLP\\
	ResNetMLP &$\gets$ &ResNet &+ &MLP &+ &MLP\\
	\bottomrule
	\end{tabular}
	\begin{flushleft}
	\justifying
		"ViT": Vision Transformer \cite{16x16WORDS}, 
		"LLaMA~II": Large Language Model Meta AI \cite{LLaMAII}, 
		"ResNet": Residual Neural Network \cite{ResNet}, 
		"Cross~T".: Cross Transformer (with Cross Attention) \cite{MultiModalTransformersSurvey}, 
		"MLP": Multi Layer Perceptron
	\end{flushleft}
	\label{tab:archs}
\end{table}

\subsection{Chest X-Ray + Clinical Report}

\subsubsection{Task: Classification}
For the disease multi-class classification we used the ViT\-LLaMAII, as already done by \cite{Gapp_Multimodal_Medical_Disease_Classification}, and additionally trained a ResNet\-LLAMAII, a ViT\-MLP and a ResNet\-MLP. The task's goal is to detect diseases regarding the lung using image--text pairs.
Following \cite{Gapp_Multimodal_Medical_Disease_Classification,TransCheX}, the 3,677 image--text data-pairs were split with 87:3:10 into training (3,199), validation (101) and testing (377) datasets.

\subsubsection{Dataset}
\begin{figure}[h!]
	\centering
	\begin{minipage}[t]{.4\textwidth}
		\begin{figure}[H]
			\includegraphics[width=0.9\textwidth]{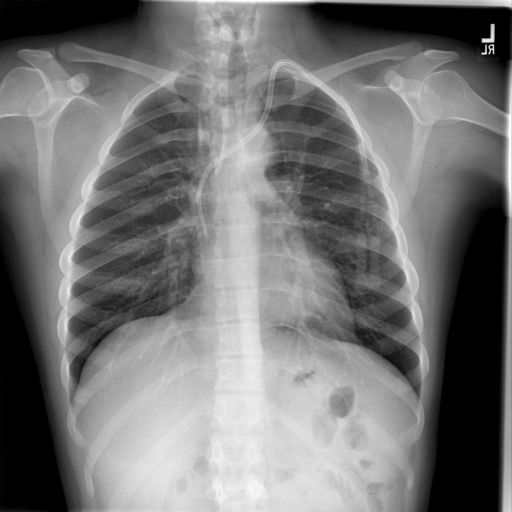}
		\end{figure}
	\end{minipage}
\begin{minipage}[t]{.6\textwidth}
	\begin{figure}[H]
		\fbox{\parbox{\textwidth-6pt}{
				\texttt{FINDINGS : The heart is normal in size. The mediastinum is within normal limits. Pectus deformity is noted. Left IJ dual-lumen catheter is visualized without \colorbox{orange}{pneumothorax}. The lungs are clear. IMPRESSION : No acute disease.
				}
		}}
	\end{figure}
\end{minipage}
\caption{Chest X-Ray + Clinical Report. Example item CXR1897\_IM-0581-1001. Disease: support devices. Orange words (labels) removed during the preprocessing step.}%
\label{fig:exampleChestXRay}
\end{figure}

The image--text dataset from Open~I \cite{OpenI_dataset} contains 2D Chest X-Rays together with a clinical report for each patient (see \cref{fig:exampleChestXRay}). We used data from 3,677 patients the same way as done in \cite{TransCheX}. In addition, we removed the class labels from the text. The 14 target classes include twelve diseases regarding the chest, especially the lungs (i.e. atelectasis, cardiomegaly, consolidation, edema, enlarged cardiomediastinum, fracture, lung lesion, lung opacity, pleural effusion, pleural other, pneumonia, pneumothorax), one for support devices and one for no finding.

\subsection{BRSET}

\subsubsection{Task: Classification}
For this multi class classification problem we trained a ResNet\-MLP and a ViT\-MLP model. The goal is to detect multiple diseases by training a network with image--tabular data.
We used 13,012 data-pairs for training and 3,254 for testing (i.e. a split of 80:20) in our trainings routine.

\subsubsection{Dataset}
\begin{figure}[h!]
	\centering
	\subfigure[source img (2390$\times$1880)]{\label{fig:sourceBRSET}\includegraphics[width=0.3802\textwidth]{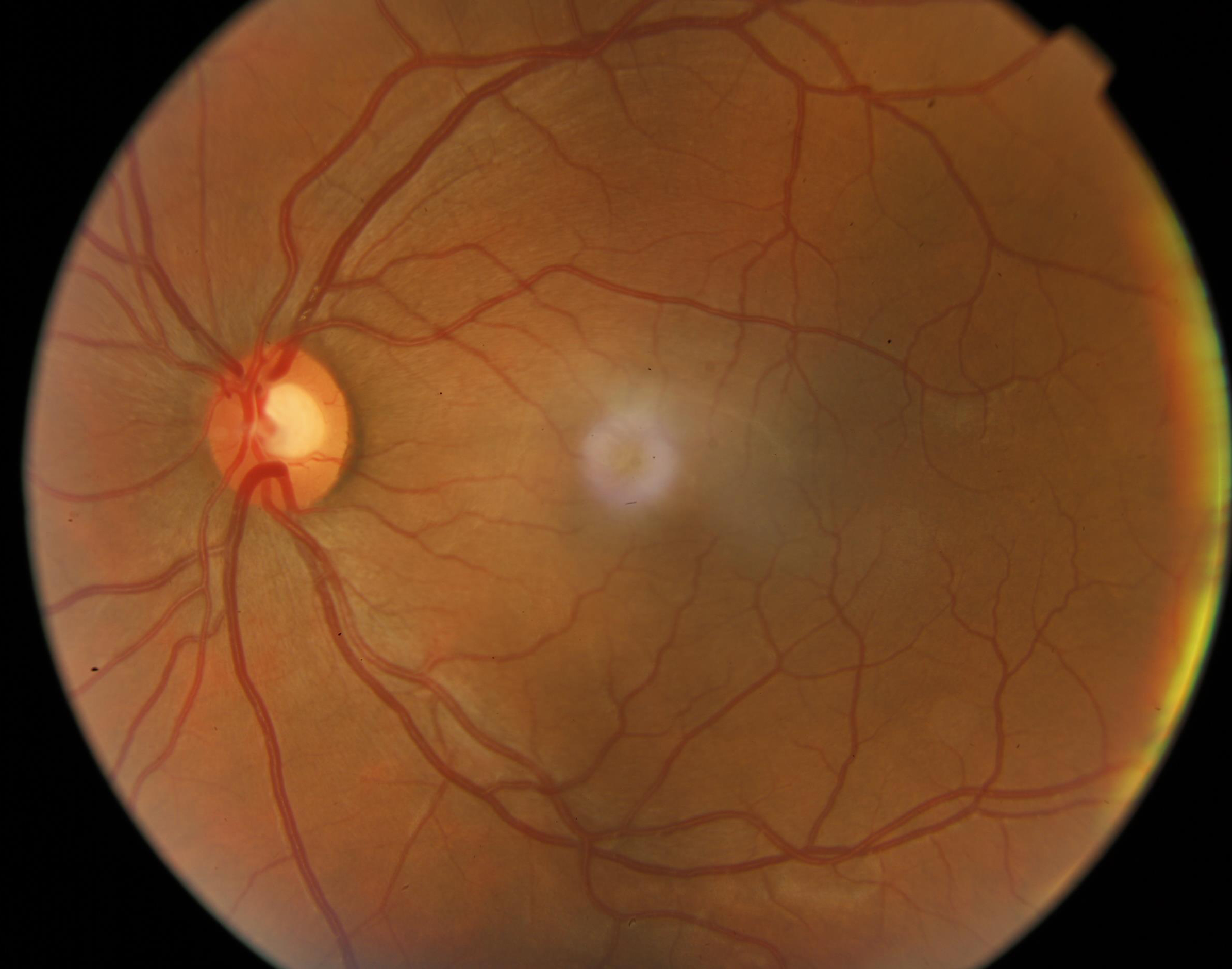}\centering}
	\hspace{2mm}
	\subfigure[transf. img (960$\times$1120)]{\label{fig:transformed}\includegraphics[width=0.35\textwidth]{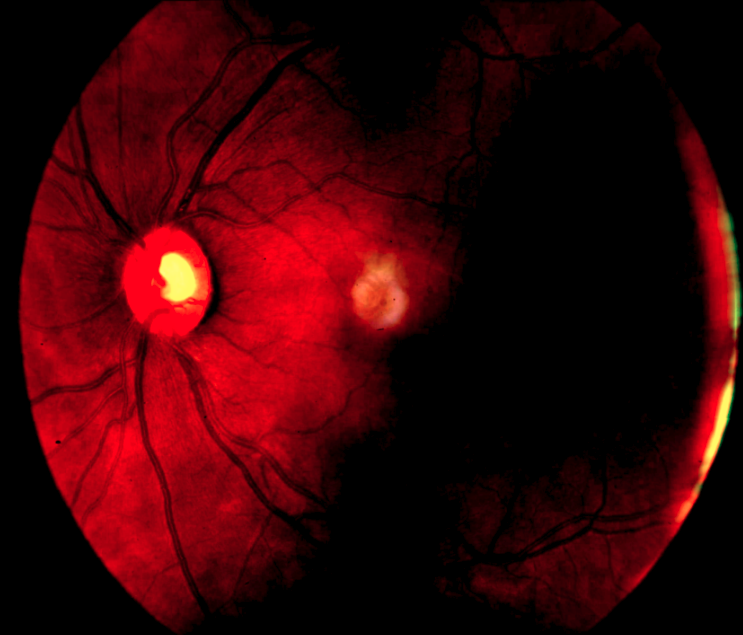}\centering}
	\caption{BRSET. Image img01468. Preprocessing for trainings routine. The source images were normalized with mean = [0.485, 0.456, 0.406] and std = [0.229, 0.224, 0.225]. Note that black parts in the transformed image inside the eye are still distinguishable by the model.} 
	\label{fig:souretransformedBRSET}
\end{figure}

The Brazilian Multilabel Ophthalmological Dataset (BRSET) \cite{BRSET} includes 2D color fundus retinal photos (see \cref{fig:sourceBRSET}) as well as patient specific data in tabular form as presented in \cref{tab:TabularDataBRSET}. The authors provide 16,266 image--tabular data-pairs from 8,524 patients. The 14 target classes can be used for multimodal disease classification. The classes include 13 diseases (i.e. diabetic retinopathy, macular edema, scar, nevus, amd, vascular occlusion, hypertensive retinopathy, drusens, hemorrhage, retinal detachment, myoptic fundus, increased cup disc, other) and one class indicating no finding.

\begin{table}[h!]
	\caption{Tabular data in BRSET.}
	\centering
	\begin{tabular}{ll}
		\toprule
		\multicolumn{1}{c}{type}  & \multicolumn{1}{l}{description}           \\
		\cmidrule(){1-2}
		patient age:  & age of patient in years\\
		comorbidities: & free text of self-referred clinical antecedents\\
		diabetes time: & self-referred time of diabetes diagnosis in years\\
		insulin use: & self-referred use of insulin (yes or no)\\
		patient sex: & enumerated values: 1 for male and 2 for female\\
		exam eye: & enumerated values: 1 for the right eye and 2 for the left eye\\
		diabetes: & diabetes diagnosis (yes or no)\\
		\bottomrule
	\end{tabular}
	\label{tab:TabularDataBRSET}
\end{table}

\subsection{Hecktor 22}

\subsubsection{Task: Regression}
For the regression we used the ResNet\-MLP and the ViT\-MLP models, with Rectified Linear Units (ReLUs) as activation functions here. The task's primary goal is to predict the RFS, PFS time of the patients using image--tabular data.
For the regression training we used data from 355 patients for training and 89 patients for testing, i.e. a split of approximately 80:20. 

\subsubsection{Dataset}
The head and neck tumor segmentation training dataset contains data from 524 patients. For each patient 3D CT, 3D PET images and segmentation masks of the extracted tumors are provided together with clinical information in the form of tabular data as presented in \cref{tab:TabularDataHECKTOR}. For the RFS (Relapse Free Survival) time prediction task, i.e. a regression problem, labels (0,1), indicating the occurrence of relapse and the RFS times (for label 0) and the PFS (Progressive Free Survival) times (label 1) in days are made available for 488 patients.
Due to some incomplete data and non-fitting segmentation masks, we finally could use image--tabular data pairs (CT segmentation mask + tabular data) of 444 patients. 
\begin{table}[h!]
	\caption{Tabular data in Hecktor~22.}
	\centering
	\begin{tabular}{ll}
		\toprule
		\multicolumn{1}{c}{type}  & \multicolumn{1}{l}{description}           \\
		\cmidrule(){1-2}
		gender:  &male (M), female (F) \\
		age: &patient age in years\\
		weight: &patient weight in kg\\
		tobacco: &smoker (1), non-smoker (0)\\
		alcohol: &drinks regularly (1), non-alcoholic (0)\\
		performance status: & 1 or 0\\
		HPV status: & positive (1), negative (0)\\
		surgery: & had a surgery (1), no surgery (0)\\
		chemotherapy: & gets chemotherapy (1), no chemotherapy (0)\\
		\bottomrule
	\end{tabular}
	\label{tab:TabularDataHECKTOR}
\end{table}

\subsection{Performance Results for Multimodal Medical Taks}
 Detailed performance results are provided in \cref{tab:performanceChestXRay} for Chest X-Ray, in \cref{tab:performanceBRSET} for BRSET and in \cref{tab:performancehecktor} for Hecktor~22.
 
\begin{table}[h!]	
	\caption{Performance AUC on image--text pair classification task, i.e. Chest X-Ray + clinical report, for multimodal, vision and clinical models using the testing dataset.}
	\centering
	\begin{tabular}{llc}
		\toprule
		modality &model &mean AUC\\
		\cmidrule{1-3}
		multimodal &ViTLLaMAII &$\mathbf{0.966}$\\
		multimodal &ResNetLLaMAII &$0.927$\\
		multimodal &ResNetMLP &$0.892$\\
		multimodal &ViTMLP &$0.904$\\
		vision &ViT &$0.629$\\
		vision &ResNet &$0.677$\\
		clinical &LLaMAII &$0.941$\\
		clinical &MLP &$0.921$\\
		\bottomrule
	\end{tabular}
	\label{tab:performanceChestXRay}
\end{table}

\begin{table}[h!]	
	\caption{Performance AUC on the image--tabular pair classification task BRSET, for multimodal, vision and clinical models using the testing dataset.}
	\centering
	\begin{tabular}{llc}
		\toprule
		modality &model &mean AUC\\
		\cmidrule{1-3}
		multimodal &ResNetMLP & $\mathbf{0.907}$ \\
		multimodal &ViTMLP &$0.798$ \\
		vision &ResNet &$0.899$ \\
		vision &ViT &$0.724$ \\
		clinical &MLP &$0.669$ \\
		\bottomrule
	\end{tabular}
	\label{tab:performanceBRSET}
\end{table}

\begin{table}[h!]
	\caption{Performance c-index on image--tabular regression task, viz. Hecktor~22, for multimodal, vision and clinical models on the testing dataset. Here the results are for the RFS prediction, i.e. the observed class is label~1: relapse.
	}
	\centering
	\begin{tabular}{llc}
		\toprule
		modality &model &c-index\\
		\cmidrule{1-3}
		multimodal &ResNetMLP &$0.705$ \\
		multimodal &ViTMLP &$0.605$ \\
		vision &ResNet &$\mathbf{0.709}$ \\
		vision &ViT &$0.678$ \\
		clinical &MLP &$0.572$ \\
		\bottomrule
	\end{tabular}
	\label{tab:performancehecktor}
\end{table}

\begin{table}[h!]
	\caption{Modality contribution specific for architecture and dataset. Entries quantify $m_0$ to $m_1$, viz. vision~:~text (Chest X-Ray) and vision~:~tabular (BRSET, Hecktor~22). The computation was done on the testing datasets.}
	\centering
	\begin{tabular}{lccc}
		\toprule
		\multicolumn{1}{c}{Model} & \multicolumn{1}{c}{Chest X-Ray}& \multicolumn{1}{c}{BRSET} & \multicolumn{1}{c}{Hecktor~22}\\
		\cmidrule(r){1-1} \cmidrule(r){2-2} \cmidrule{3-3} \cmidrule(l){4-4}
		ResNetLLaMAII
		& $0.18 : 0.82$ 
		&  
		& \\
		ViTLLaMAII
		& $0.13 : 0.87$ 
		&  
		& \\
		ResNetMLP
		& $0.59 : 0.41$  
		& $0.92 : 0.08$ 
		& $0.65 : 0.35$ 
		\\
		ViTMLP
		& $0.02 : 0.98$ 
		& $0.67 : 0.33 $ 
		& $0.00 : 1.00$\\ 
		\bottomrule
	\end{tabular}
	\label{tab:ModalityContrib}
\end{table}

\section{Interpretability Results} \label{sec:Experiments}
For both classification tasks and the regression problem we present our quantitative metric $m_i$ to measure the modality contributions. In \cref{tab:ModalityContrib} the results are summarized. For datasets BRSET and Hecktor~22 we also present the modality specific importance $mp_i^l$.

\subsection{Multiclass Classification with Chest X-Ray + Clinical Report}

\begin{figure}[h!]
	\centering
	\includegraphics[width=\textwidth]{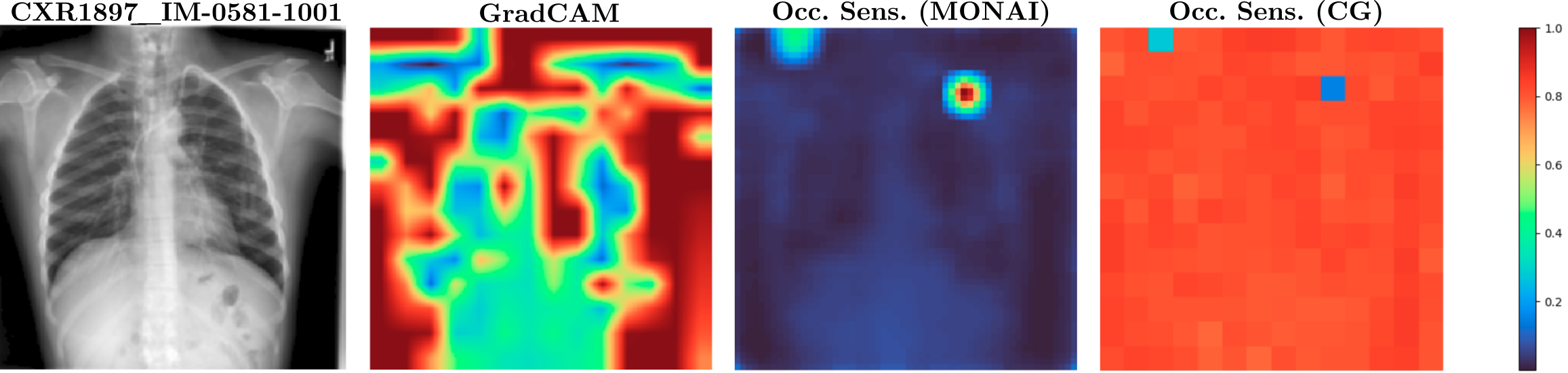}
	\includegraphics[width=\textwidth]{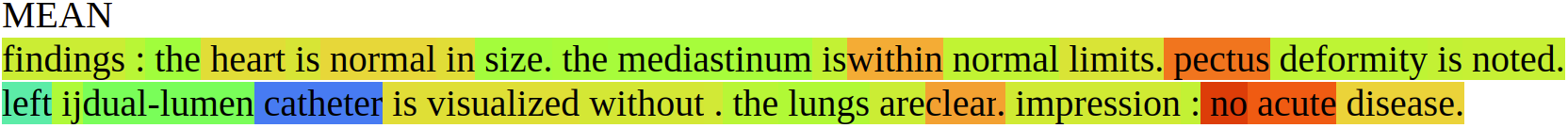}
	\includegraphics[width=\textwidth]{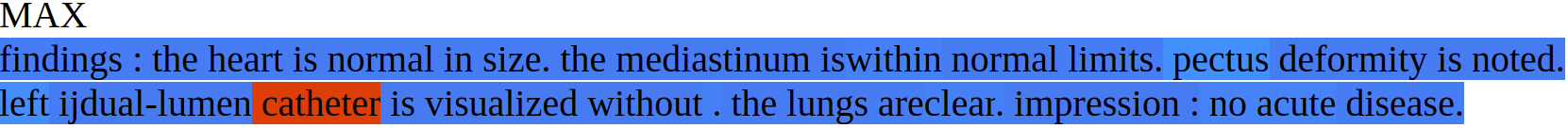}
	\caption{CXR1897\_IM-0581-1001: Correctly predicted disease: support devices. Modality contribution vision~:~text = 0.24~:~0.76. Model: ViTLLAMA~II. From blue to red the contribution (low to high) from a single patch (vision) or word (text) to the task is highlighted. Top, left to right: source image, GradCAM, class specific Occlusion Sensitivity for class support devices (MONAI), Occlusion Sensitivity averaged over all classes (CG, i.e. \emph{ours}). The red patch in the upper right area in image Occ. sens. (MONAI) has the highest contribution to the class support devices. The same area is colored blue in image Occ. sens. (CG), as this patch has the lowest average contribution to all classes.  Bottom: Text. MEAN: The words \emph{no} and \emph{acute} have the highest average contribution, \emph{catheter} has the lowest. 
	MAX: \emph{catheter} has the highest contribution to one class: support devices.}
	\label{fig:ChestXRay_saliency}
\end{figure}

In \cref{tab:ModalityContrib}, second column, the $m_i$ results are printed for four different multimodal nets. The computations were done with $h_i = \lbrace$vision: 196, text: \emph{number} \emph{of} \emph{words} \emph{in} \emph{report}$\rbrace$. The visualization results and analyses are presented in \cref{fig:ChestXRay_saliency}.

\begin{figure}[h!] 
	\includegraphics[width=\textwidth]{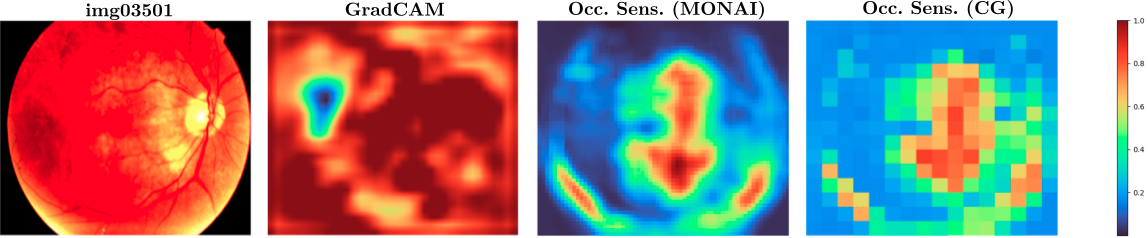}
	\includegraphics[width=.6\textwidth]{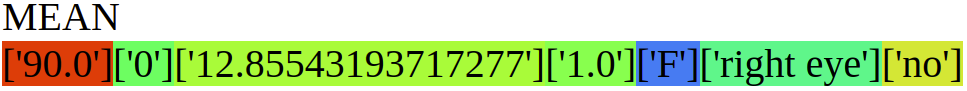}\\
	\includegraphics[width=.6\textwidth]{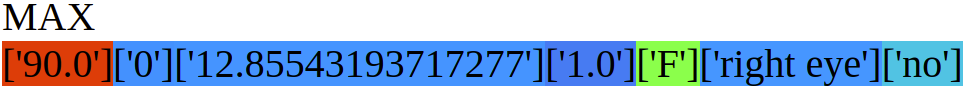}
	\caption{img03501: Correctly predicted disease: drusens. Modality contribution vision~:~tabular~= 0.95~:~0.05. Model: ResNet\-MLP. Importance (low to high) is colored from blue to red. Top, left to right: source image, GradCAM, class specific Occlusion Sensitivity for class drusens (MONAI), Occlusion Sensitivity averaged over all classes (CG, i.e. \emph{ours}). Bottom: tabular data with attributes patient age, comorbidities, diabetes time, insulin use, patient sex, exam eye, diabetes from left to right. MEAN: The patient's age has the highest contribution, patient sex the lowest in average. MAX: patient's age is the most significant attribute for one class: drusens.}
	\label{fig:BRSET_saliency}
\end{figure}

\subsection{Multiclass Classification with BRSET}

The third column of \cref{tab:ModalityContrib} shows the $m_i$ results for the two trained nets. Modality Contribution ratios (vision : tabular) are $0.92 : 0.08$ for ResNet\-MLP and $0.67 : 0.33$ for ViT\-MLP. The computations were done with $h_i = \lbrace$vision: 240, tabular: 7$\rbrace$.
The visualization results are presented in \cref{fig:BRSET_saliency} together with explanations. The token importance for tabular data is presented in \cref{tab:mpi_BRSET}.
\begin{table}[h!]
	\caption{Modality Contribution $m_i^l$ and $mp_i^l$ with $l = 0,\dots,h_i-1$ and $h_i=7$ for the tabular data in the BRSET classification problem. The three most important attributes per $mp_i^l$-column are highlighted in bold.} 
	\centering
	\begin{tabular}{lccccc}
		\toprule
		&\multicolumn{2}{c}{ResNetMLP} & \multicolumn{2}{c}{ViTMLP}\\
		\cmidrule(r){2-3}
		\cmidrule{4-5}
		&\multicolumn{1}{c}{$m_i^l$} & \multicolumn{1}{c}{$mp_i^l$} &\multicolumn{1}{c}{$m_i^l$} & \multicolumn{1}{c}{$mp_i^l$}& \multicolumn{1}{c}{$l$}\\
		\cmidrule{1-6}
		\textbf{patient age}:  &$0.024$ &$\mathbf{0.289}$ &$0.081$&$\mathbf{0.245}$ &$0$\\
		comorbidities: &$0.008$ &$0.093$ &$0.025$&$0.076$ &$1$\\
		\textbf{diabetes time}: &$0.011$ &$\mathbf{0.133}$ &$0.063$&$\mathbf{0.192}$ &$2$\\
		insulin use: &$0.009$ &$0.110$ &$0.047$&$0.143$ &$3$\\
		patient sex: &$0.010$ &$0.124$ &$0.018$&$0.056$ &$4$\\
		exam eye: &$0.006$ &$0.091$ &$0.033$&$0.100$ &$5$\\
		\textbf{diabetes}: &$0.013$ &$\mathbf{0.160}$ &$0.062$&$\mathbf{0.188}$ &$6$\\
		\cmidrule{1-6}
		sum: &$0.081$ &$1.000$ &$0.329$&$1.000$ &\\
		\bottomrule
	\end{tabular}
	\label{tab:mpi_BRSET}
\end{table}

\subsection{Regression -- RFS Prediction with Hecktor~22}

For the computation of the modality contribution for the RFS prediction task we have chosen $h_i = \lbrace$vision: 8, tabular: 9$\rbrace$. The big mask, i.e. half of image size in each dimension ($2^{-3} = 8^{-1}$) for vision is necessary to occlude enough from the segmented tumor.
With ResNet\-MLP we have computed a modality contribution vision~:~tabular~= 0.65~:~0.35 (see \cref{tab:ModalityContrib}: fourth column). For ViT\-MLP we have vision~:~tabular~= 0.0~:~1.0 (unimodal collapse).
\cref{tab:mpi_hecktor22} shows the results for the token importance for the tabular data.

\begin{table}[h!]
	\caption{Modality Contribution $m_i^l$ and $mp_i^l$ with $l = 0,\dots,h_i-1$ and $h_i = 9$ for tabular data in Hecktor~22 regression problem. Architecture: ResNet\-MLP. The three most important attributes are highlighted in bold.}
\centering
\begin{tabular}{lccc}
	\toprule
	&\multicolumn{1}{c}{$m_i^l$} & \multicolumn{1}{c}{$mp_i^l$} & \multicolumn{1}{c}{$l$}\\
	\cmidrule{1-4}
	gender &0.054 
	& 0.156
	&0
	\\
	\textbf{age} & 0.069
	& \textbf{0.197}
	&1
	\\
	\textbf{weight} & 0.068
	& \textbf{0.196}
	&2
	\\
	tobacco & 0.002
	& 0.007
	&3
	\\
	alcohol & 0.034
	& 0.097
	&4
	\\
	performance status & 0.003
	& 0.008
	&5
	\\
	\textbf{HPV status} & 0.088
	& \textbf{0.254}
	&6
	\\
	surgery & 0.006
	& 0.018
	&7
	\\
	chemotherapy & 0.023
	&0.067
	&8
	\\
	\cmidrule{1-4}
	sum & 0.347 & 1.000 \\
	\bottomrule
\end{tabular}
\label{tab:mpi_hecktor22}
\end{table}

\section{Discussion}\label{sec:Discussion}
 
For the classification task with the image--text dataset from \cite{OpenI_dataset}, viz. Chest X-Ray + clinical report, we computed the modality contribution $m_i$ between vision and text. Our results for this dataset confirm the behavior found in \cite{haouhat2023modalityinfluencemultimodalmachine}, that text is the most important modality in most multimodal datasets containing text. 

Results also show that our multimodal models that contain a Residual Neural Network (ResNet) \cite{ResNet} instead of a Vision Transformer (ViT) \cite{16x16WORDS} have a bigger ratio between the vision and text modality contribution for our tasks.

Unimodal collapses occured in two cases: (1) ViTMLP for Chest X-Ray + clinical report and (2) ViT\-MLP for Hecktor~22. ResNet\-MLP for BRSET has a tendency to collapse with 92\% vision to only 8\% tabular contribution. While in (1) the ViT\-MLP does almost not use the vision modality, it is completely ignored in (2). The ViT\-MLP only works well for the classification task with BRSET.

Another interesting observation is a potential link between our modality contribution metric and the performance of single modality trained networks. Specifically, modalities with higher estimated contributions (e.g., $m_0$ and $m_1$ in \cref{tab:ModalityContrib}) tend to correspond to higher mean performances of the respective single-modality networks (see \cref{tab:performanceChestXRay}, \cref{tab:performanceBRSET}, \cref{tab:performancehecktor}). While this does not constitute formal validation,  it suggests that our metric may intrinsically reflect modality-specific performance.

Our occlusion based modality contribution has a user specific hyper-\-para\-meter for continuous data, i.e. $h_i$. In contrast to text or tabular data these lack natural sequencing. Although the vision modality is discrete, it also lacks natural sequencing due to dependencies of resolution and displayed content. As a consequence, the length of the occluded sequences ($\propto 1/h_i$) for continuous data, and for modalities such as vision, must be chosen small enough (i.e. big enough patches) to occlude sensitive information effectively. However, this choice is not straightforward, as the impact of mask size can vary for continuous data (e.g., vision) depending on the structure and content of individual items, as well as the specific task. We include ablation studies on the hyperparameter $h_i$ in the \appendixname, which can serve as a useful reference for selecting $h_i$ in other tasks. However, comprehensive investigations are required to establish a generalizable guideline for the selection of $h_i$ across different modalities and data types.

The masking operation, replacing inputs with other values, can introduce domain shift, potentially affecting whether output changes reflect true modality contributions or shift sensitivity. For images, we mitigate this by replacing masked regions with the mean pixel value of the same image, thereby reducing domain shift. For scalar features, the mean computed over the modality in the entire dataset is applied. Binary features are best represented with two-bit one-hot encoding, enabling masked values to be replaced with (0,~0) during occlusion.
Still, the trade-off between mask size and accurate contribution estimation remains a limitation of our method that we aim to address in future work.

\section{Summary and Outlook}\label{sec:Summary}
The availability of high-dimensional multimodal data, especially in medicine, has led to deep learning-based fusion methods. Interpretability techniques are essential to ensure model trustworthiness, a key requirement in medical applications.

Therefore, we created a metric to analyze deep learning models regarding modality preference in multimodal datasets. In contrast to existing methods \cite{gat2021perceptualscoredatamodalities,MCI_SHAP,hu2022shapeunifiedapproachevaluate,coffeBean}, our method is both fully model and performance agnostic.

We trained three different multimodal data\-sets, one image--text for classification and two image--tabular for classification and regression respectively, and evaluated the models on the testing datasets. Then we applied our new method on them. We could show that some architectures process multimodal data in a balanced way, while others tend to unimodal collapses.
Furthermore, import attributes within one modality were quantitatively highlighted.

Knowing modality contributions in multimodal medical data enables more effective model selection or redesign to prevent issues like unimodal collapse and fully harness the dataset’s potential. While our method aims to support robust integration of multimodal AI into clinical practice, it comes with limitations related to the mask size selection and the absence of explicit modeling of cross-modality interactions.
Nonetheless, we plan to address these limitations in future refinements. Code is available at {\tt{{https://github.com/ChristianGappGit/MC\_MMD}}}.

\begin{credits}
\subsubsection{\ackname}
This study is partly supported by VASCage -- Centre on Clinical Stroke Research.
\subsubsection{\discintname}
The authors have no competing interests to declare that are
relevant to the content of this article.
\end{credits}
%
%

\bibliographystyle{splncs04}
\bibliography{references}

\clearpage
\appendix
\section*{\appendixname \ -- Ablation Studies With Hyper-parameter $h$}\label{appendix}

Ablation studies were conducted on the image--text dataset 2D Chest X-Ray + clinical report with the ResNetMLP model, investigating the effect of hyper-parameter $h$. This specific dataset and model were chosen because they utilize the modalities in a largely balanced manner.
For both modalities, vision and text, we used different combinations of $h_i$ and tracked the resulting modality contribution values. In our setup we chose $h$(vision) $\in \lbrace 1,4,16,49,196,256 \rbrace$, ranging from whole-image occlusion ($224\times224$) to patch occlusion ($14\times14$), and $h$(text) $\in \lbrace1,2,4,8,16,32 \rbrace$ ranging from occluding the entire text to a single word. For instance, $h$(vision) = 49 indicates, that the occluded mask has size $32 \times32$. Since $224 / 32 = 7$ there are 7 patches along each spatial dimension, resulting in a total of $7 \cdot 7 = 49$ patches.

\begin{figure}[]
	\centering
	\includegraphics[width=0.75\textwidth]{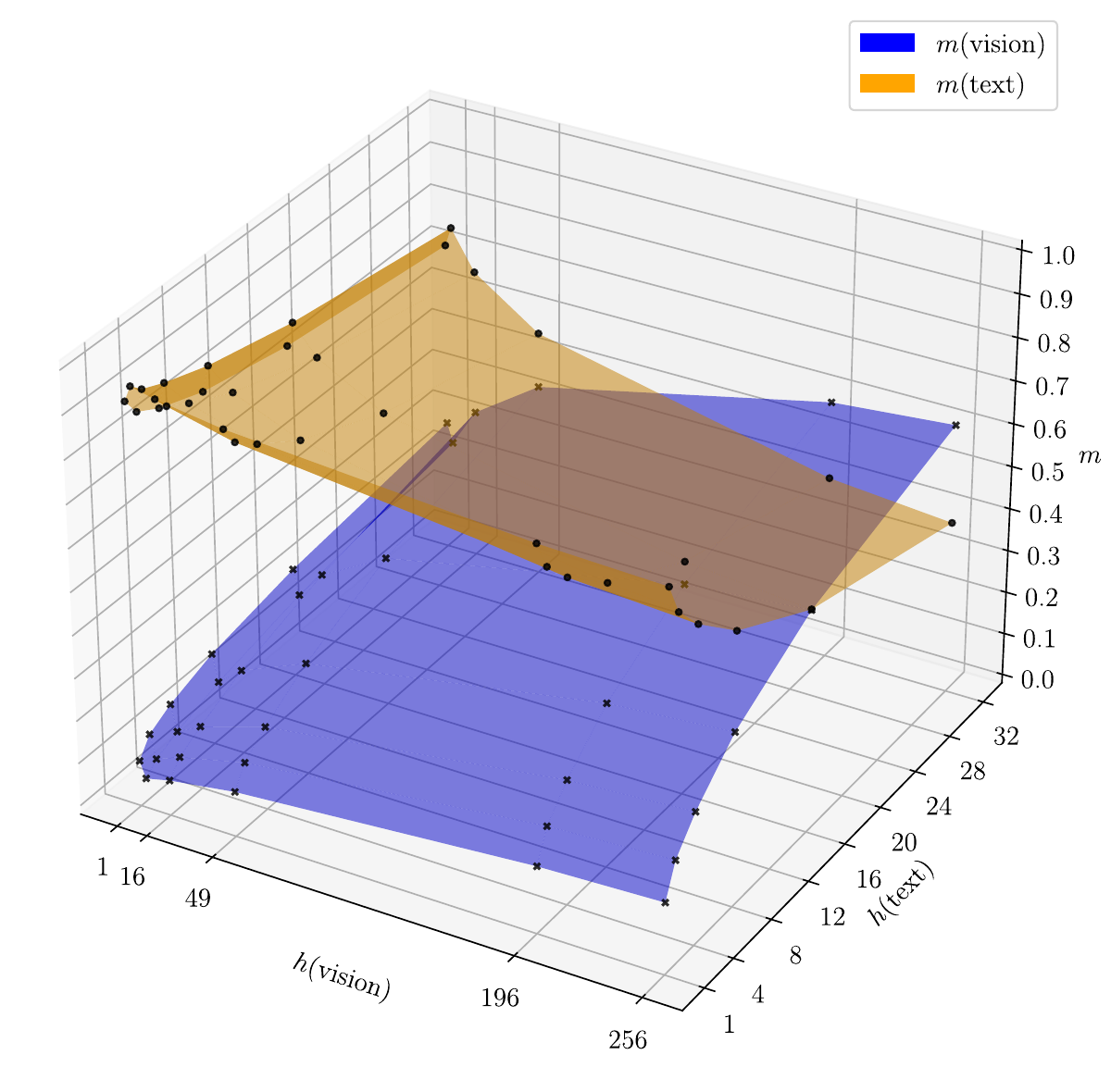}
	\caption{3D plot: $h$(vision) vs. $h$(text) vs. $m$ for both modalities, vision and text.}
	\label{fig:h_vs_m_3D}
\end{figure}

\begin{figure}[]
	\centering
	\includegraphics[width=0.32\textwidth]{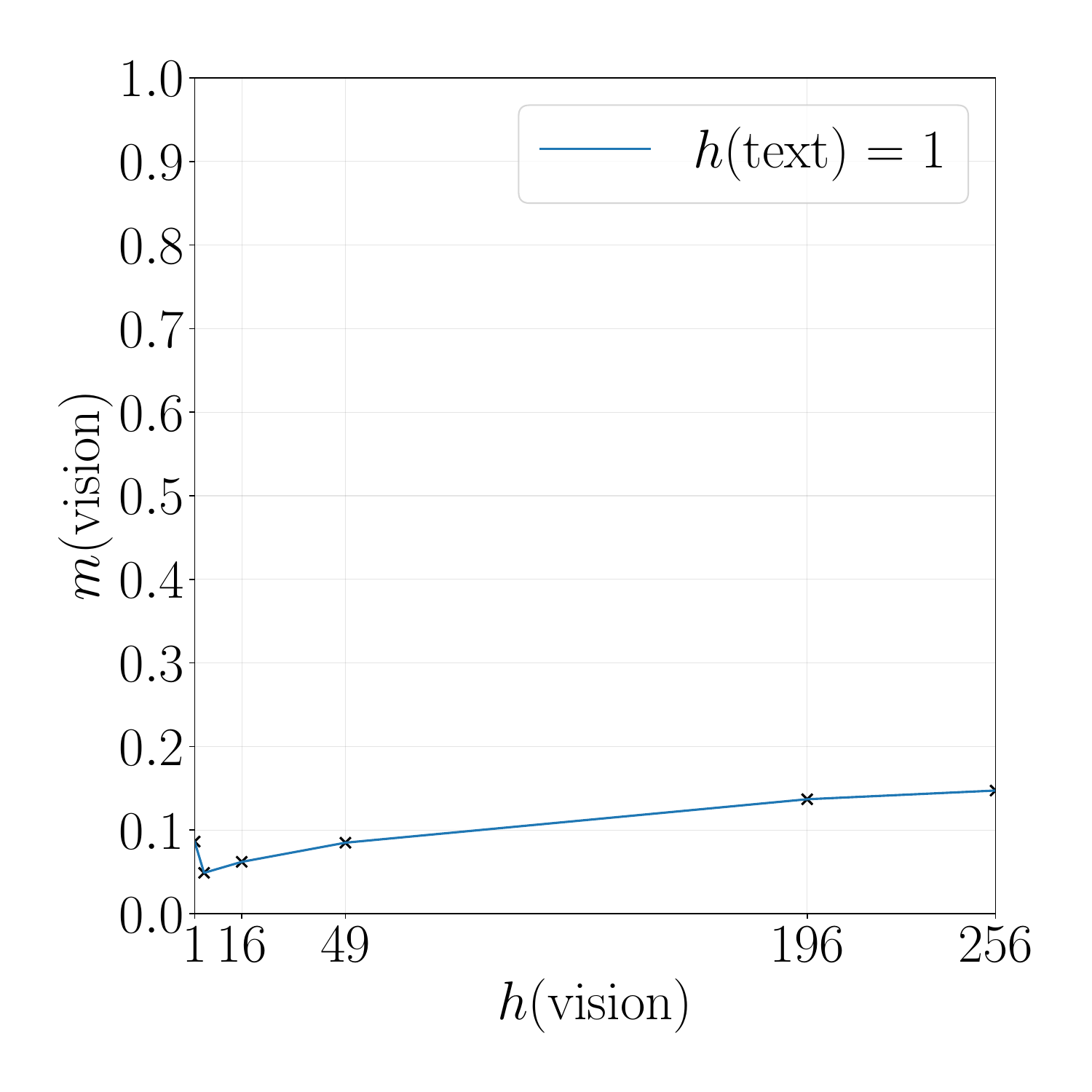}
	\includegraphics[width=0.32\textwidth]{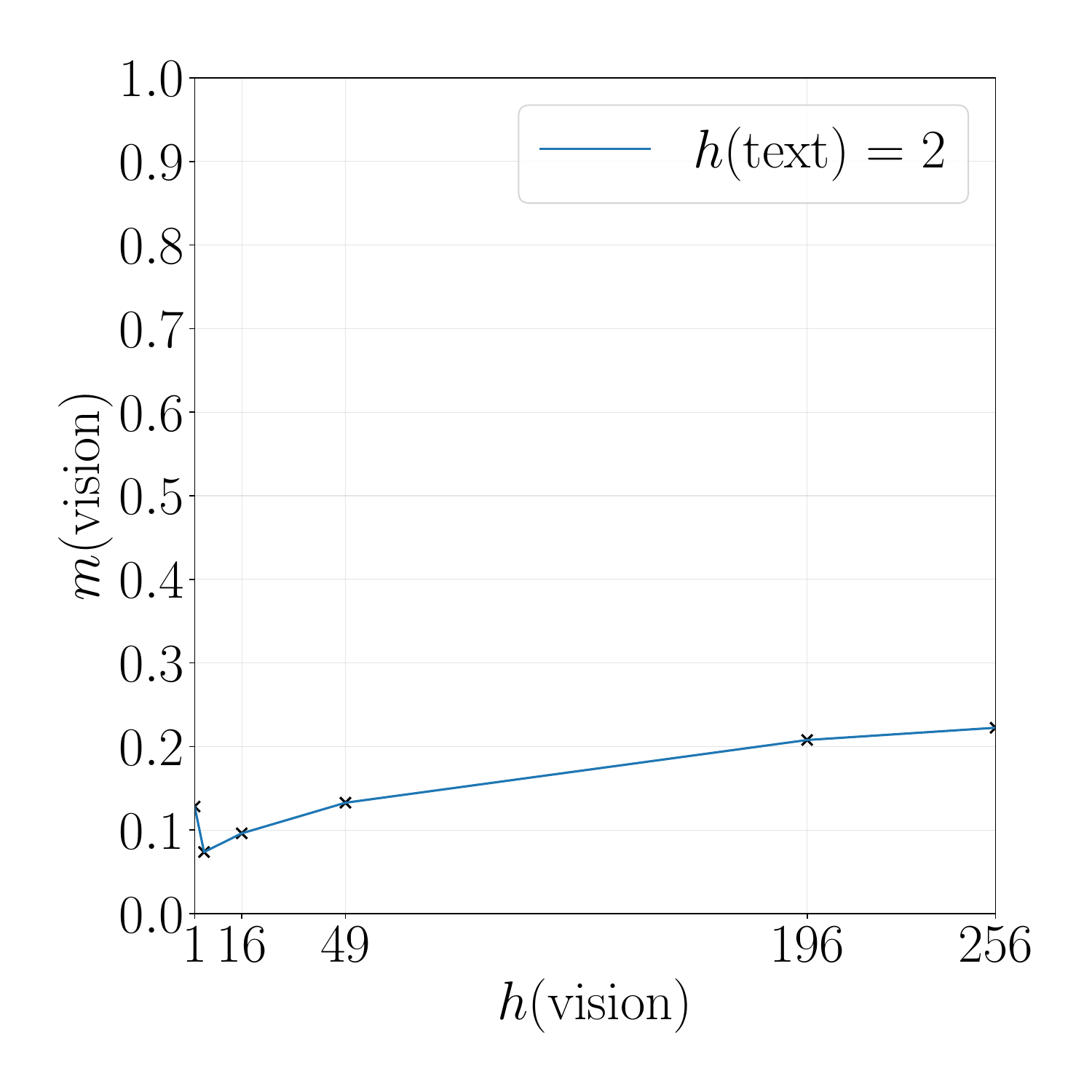}
	\includegraphics[width=0.32\textwidth]{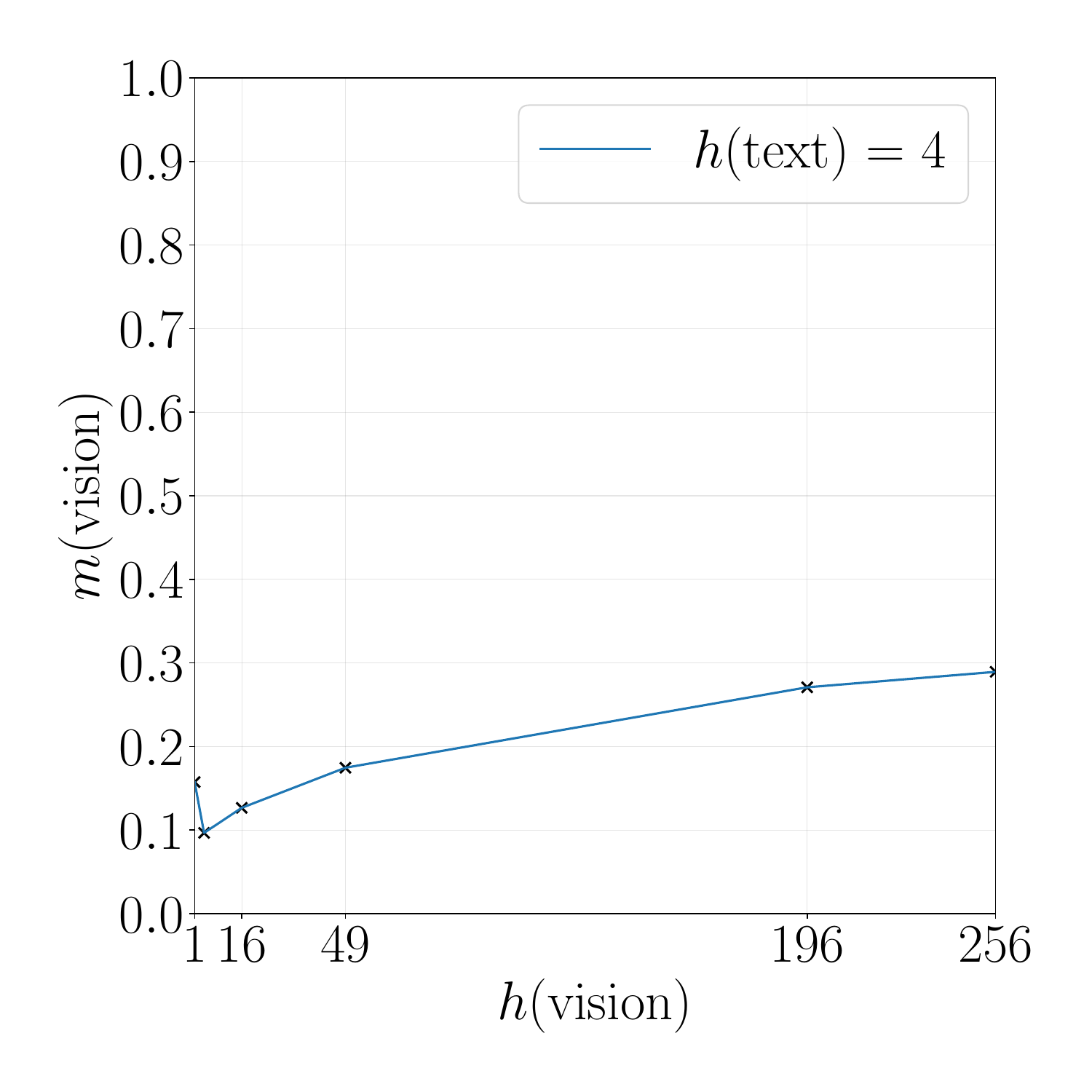}
	\includegraphics[width=0.32\textwidth]{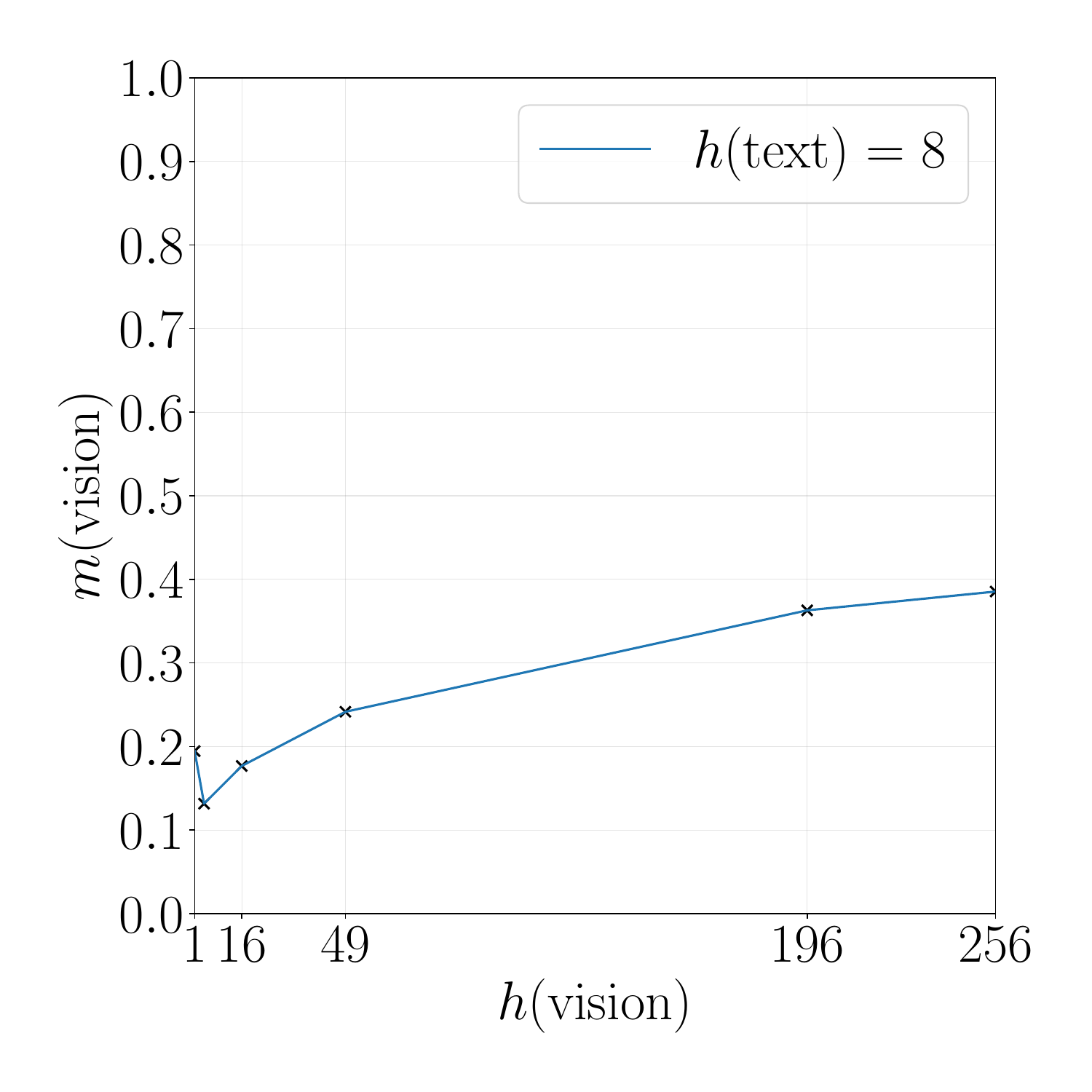}
	\includegraphics[width=0.32\textwidth]{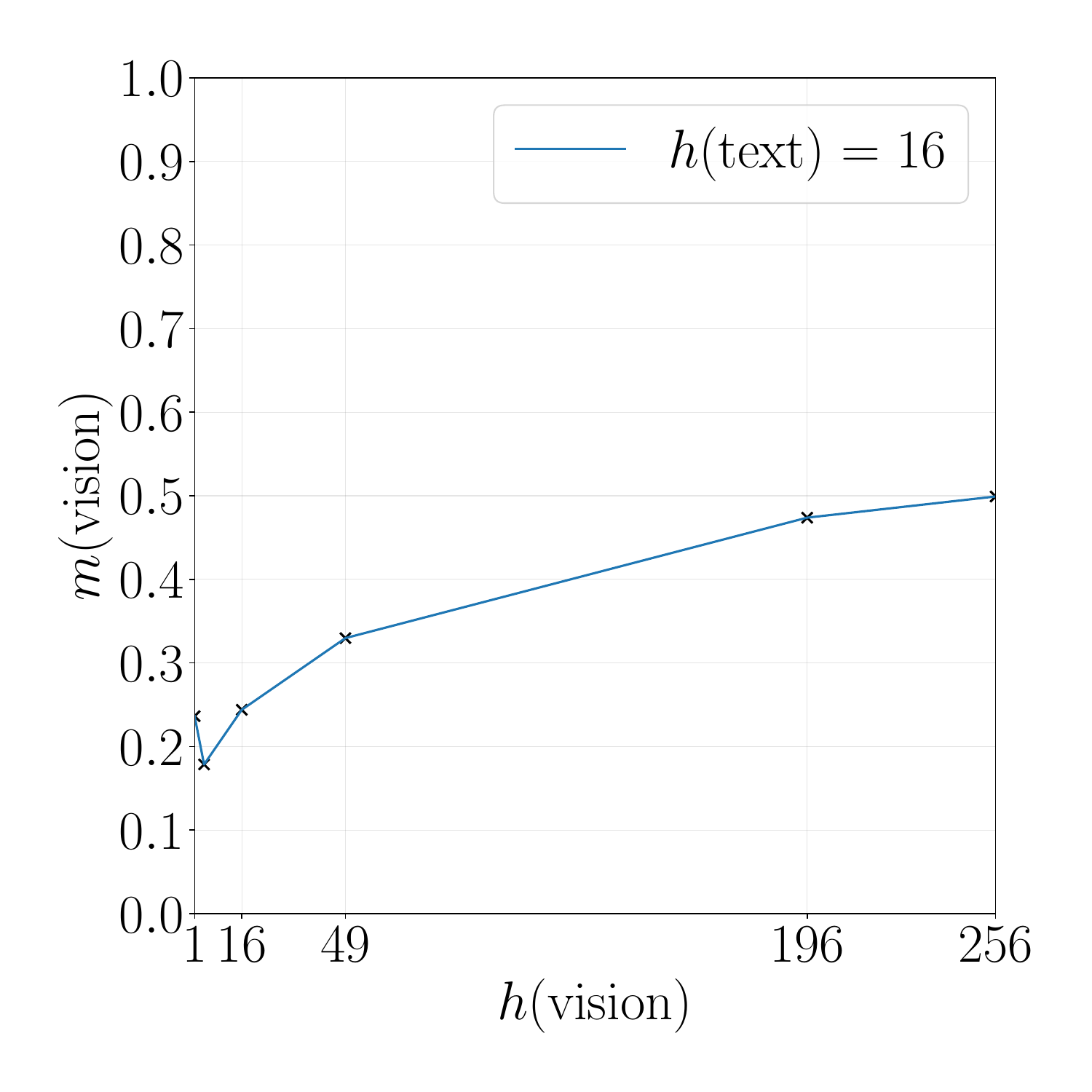}
	\includegraphics[width=0.32\textwidth]{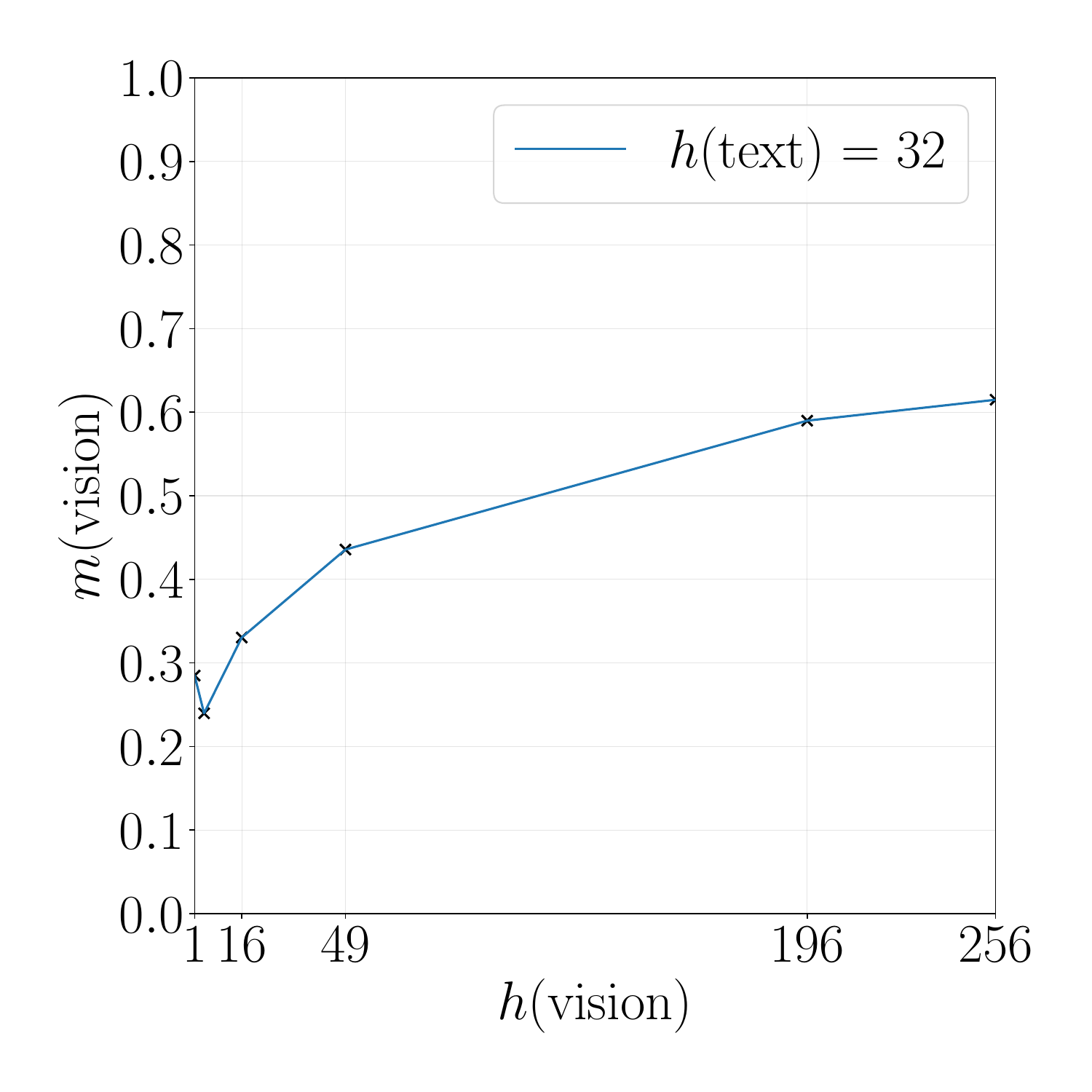}
	\caption{$h$ vs. $m$ for vision with 6 different values for $h$(text).}
	\label{fig:h_vs_m_vision}
\end{figure}

\begin{figure}[]
	\centering
	\includegraphics[width=0.32\textwidth]{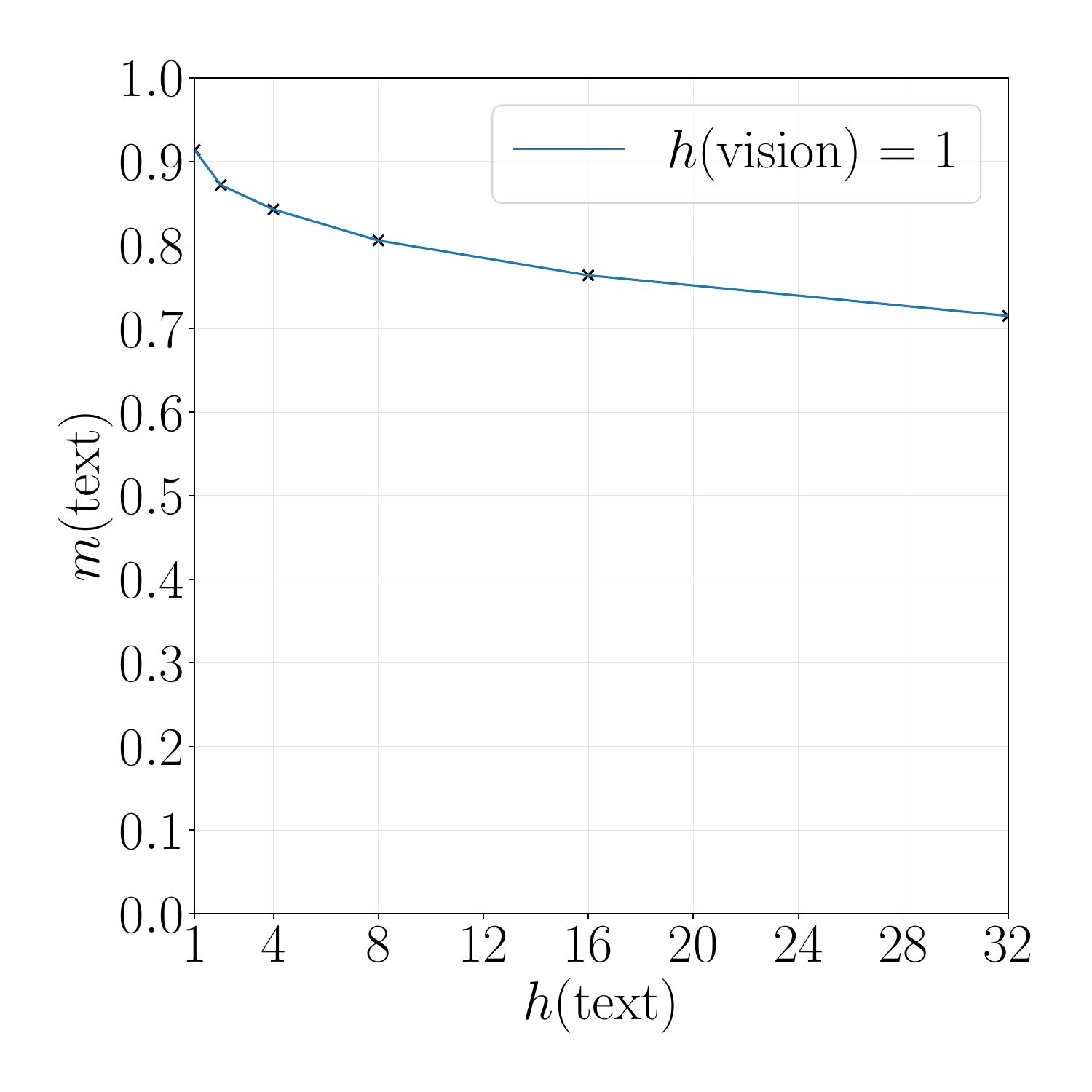}
	\includegraphics[width=0.32\textwidth]{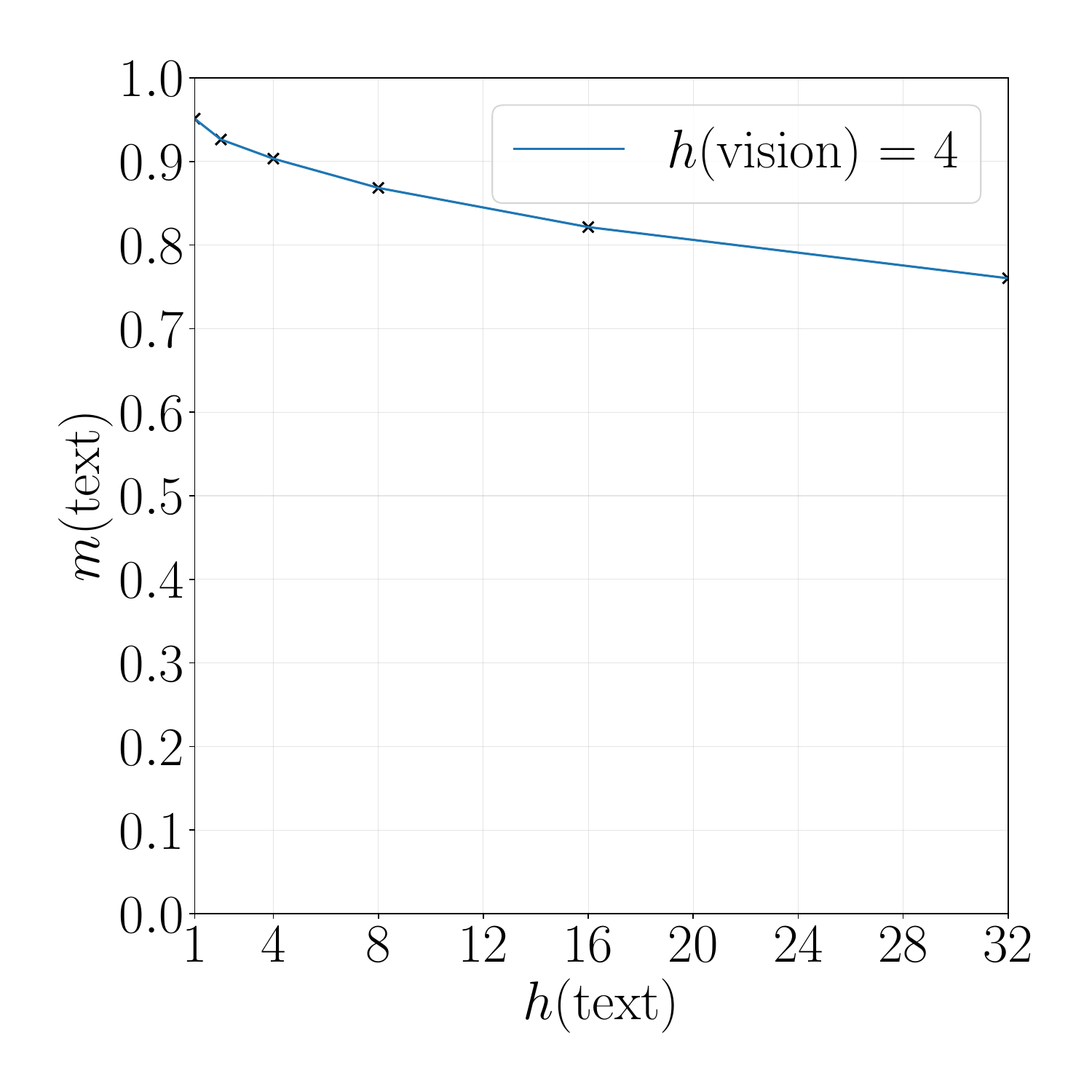}
	\includegraphics[width=0.32\textwidth]{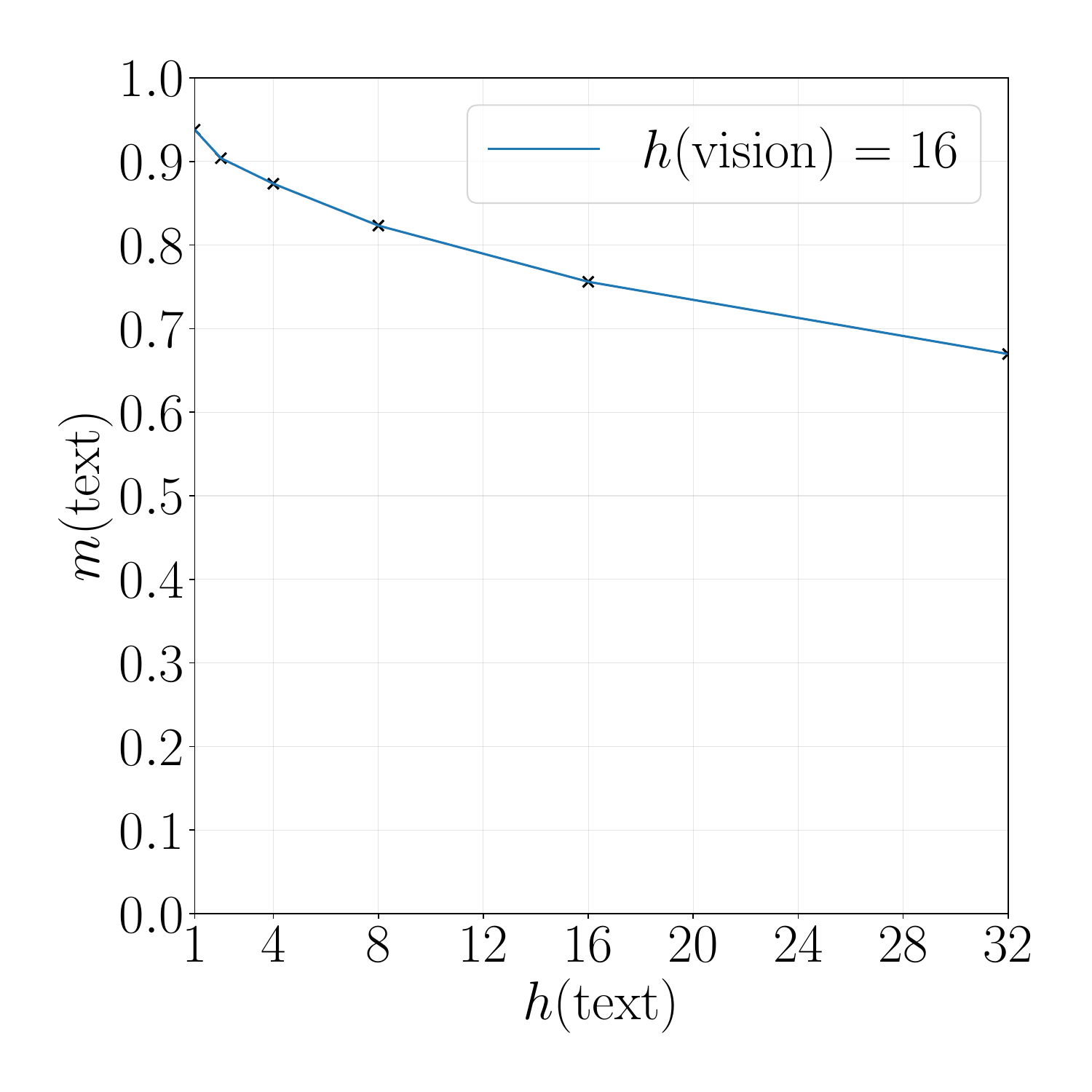}
	\includegraphics[width=0.32\textwidth]{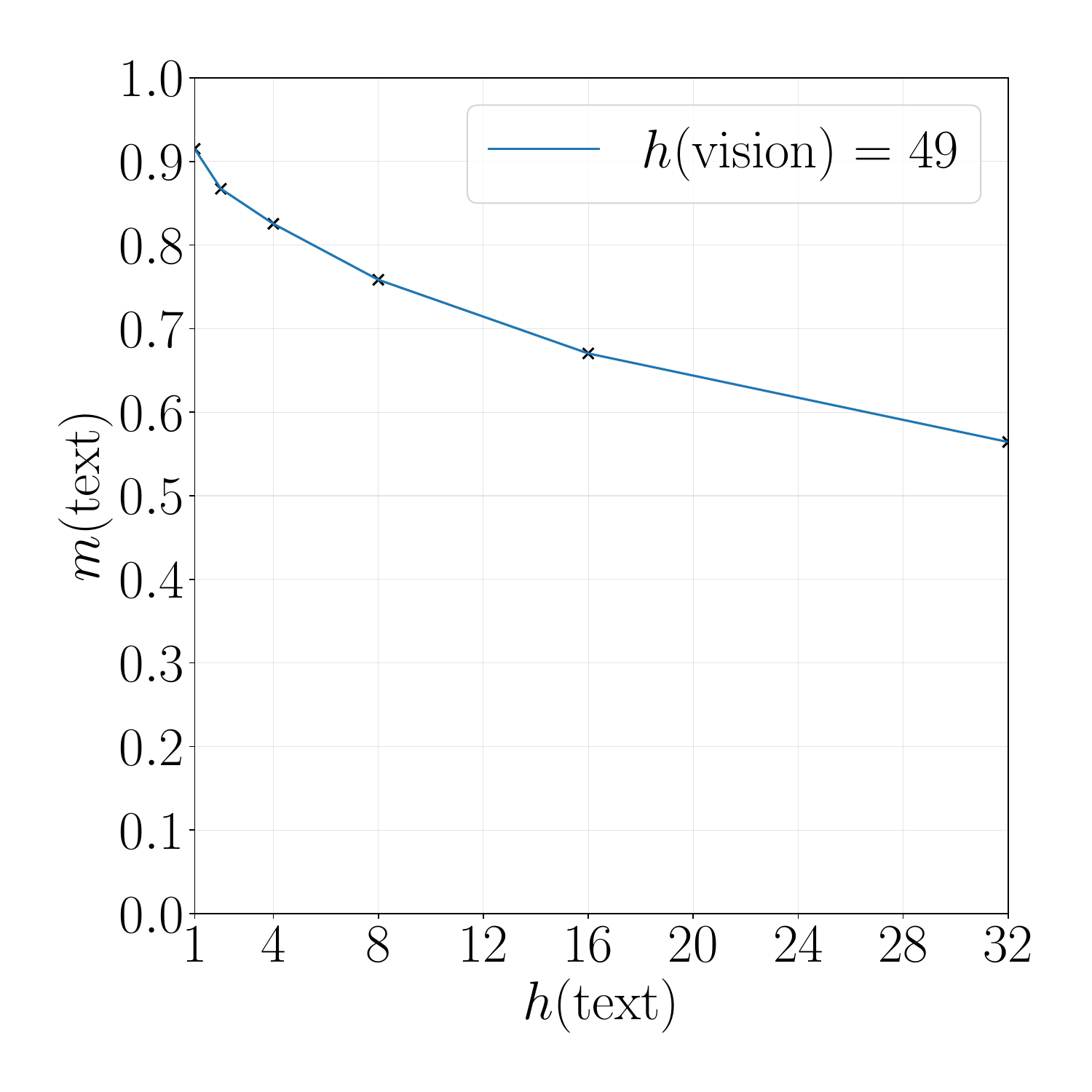}
	\includegraphics[width=0.32\textwidth]{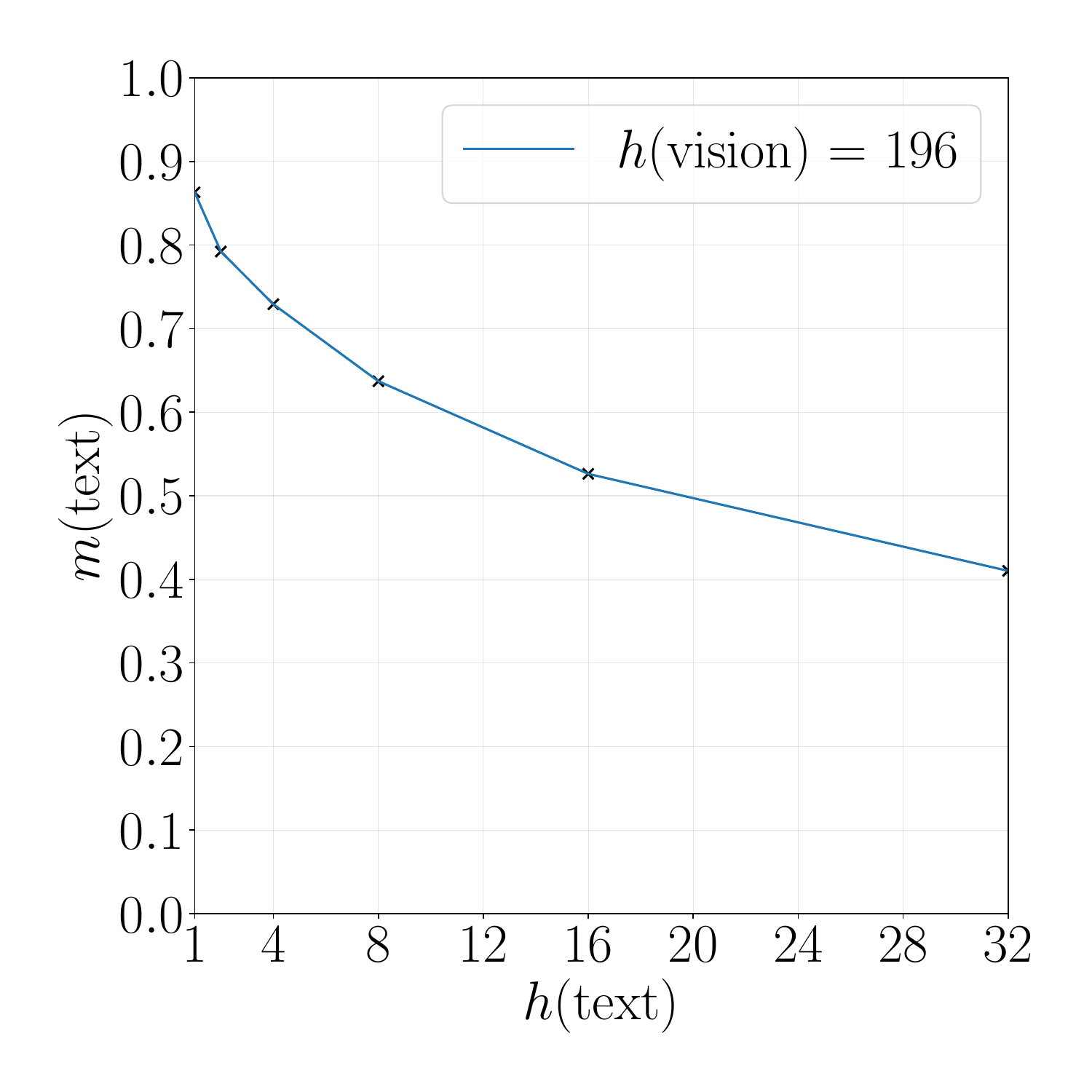}
	\includegraphics[width=0.32\textwidth]{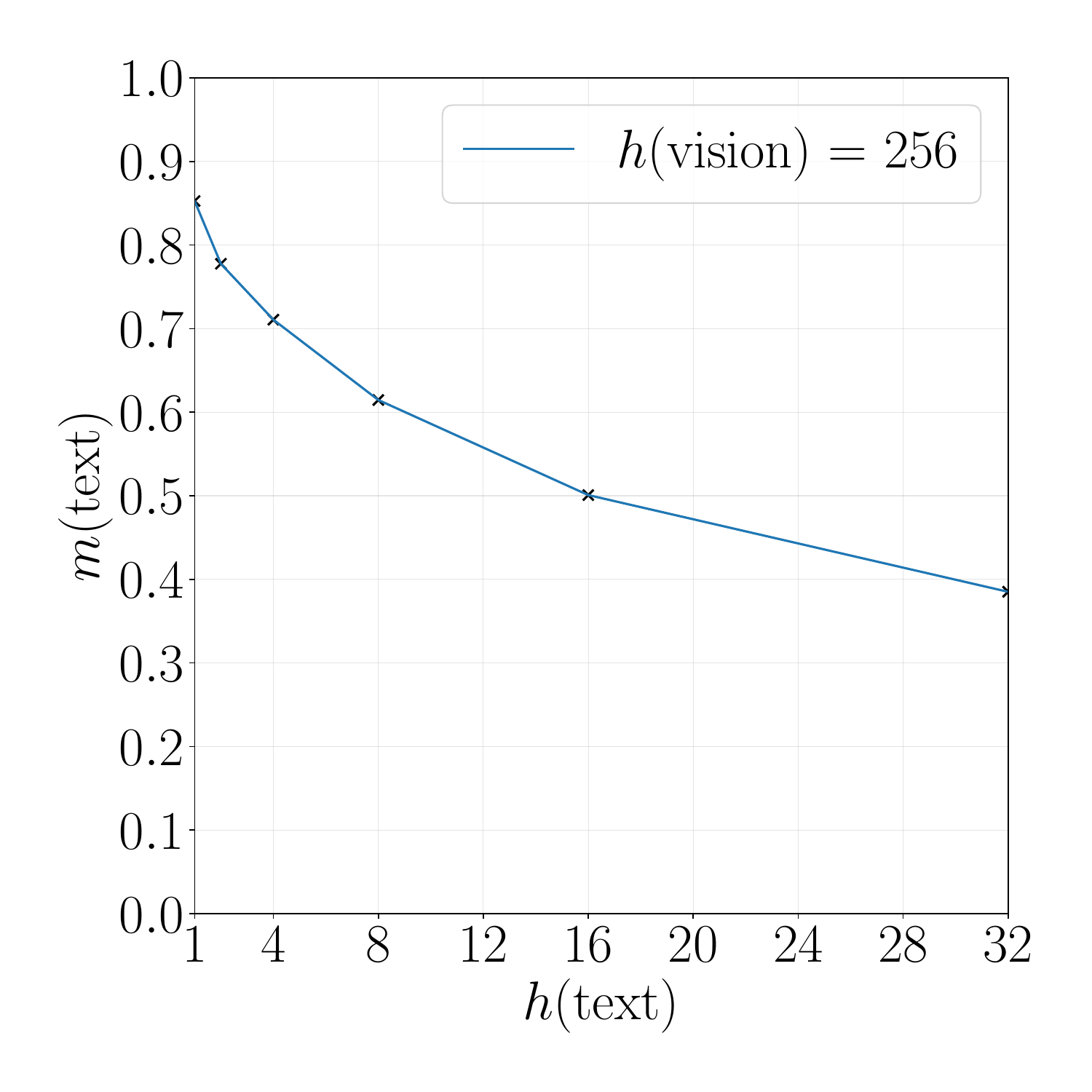}
	\caption{$h$ vs. $m$ for text with 6 different values for $h$(vision).}
	\label{fig:h_vs_m_text}
\end{figure}

Finally we plotted the resulting modality contributions $m$(vision) and $m$(text) in a 3D plot (\cref{fig:h_vs_m_3D}). The first two axes correspond to $h$(vision) and $h$(text), while the third axis shows $m$(vision) and $m$(text).
The rendered surface was generated by interpolating 36 data points.
In the \cref{fig:h_vs_m_vision,fig:h_vs_m_text} the results of the analysis for selected 2D slices are plotted, representing the variation of the modality contributions with respect to the hyper-parameters $h$(vision) and $h$(text).

\subsection*{Meaningful ranges for $h$ values}
One can determine that
\begin{equation*}
	m \text{(text)} \propto 1 / h \text{(text)}
\end{equation*}
and 
\begin{equation*}
	m \text{(vision)} \propto h \text{(vision)}, \ h \text{(text)} \, .
\end{equation*}
For this specific case, dataset and model, small values for $h$(text) (and mostly for $h$(vision)) overestimate the contribution of text, i.e. $m$(text), whereas high values for both $h$ parameters (fine occlusion) tend to overestimate $m$(vision). A reasonable and balanced choice for the hyper-parameter is $h$(vision) = $h$(text)$ = 16$, which is equal to $1/2(\text{max}(h(\text{text})))$. With this setting the contributions are $m$(vision)~:~$m$(text) = 0.24~:~0.76.

For other tasks, other datasets and models we advice setting all hyper-parameters $h_i$ with $i = 1 \dots N$ modalities to
\begin{equation*}
1/2 \, \min((\max(h_1), \max(h_2), \dots \max(h_N))) \ ,
\end{equation*}
in order to obtain an optimal estimation of the modality contributions.

\end{document}